\documentclass[fleqn,10pt]{wlscirep}
\usepackage[utf8]{inputenc}
\usepackage[T1]{fontenc}
\usepackage[english]{babel}
\usepackage{subcaption}
\usepackage{siunitx}
\usepackage{placeins}
\usepackage{helvet}
\usepackage{booktabs}
\usepackage{lineno}
\title{VisionaryVR: An Optical Simulation Tool for Evaluating and Optimizing Vision Correction Solutions in Virtual Reality}

\author[1,*]{Benedikt W. Hosp }
\author[2]{Martin Dechant }
\author[1,3]{Yannick Sauer}
\author[1]{Rajat Agarwala}
\author[1,3]{Siegfried Wahl}

\affil[1]{ZVSL, Institute for Ophthalmic Research, University of Tübingen, Maria-von-Linden Straße 6, Tübingen, Germany}
\affil[2]{UCLIC, University College London, 66 - 72 Gower Street, London, WC1E 6EA, United Kingdom}
\affil[3]{Carl Zeiss Vision International GmbH, Turnstraße 27, Aalen, Germany}

\affil[*]{benedikt.hosp@uni-tuebingen.de}

\begin{abstract}
Developing and evaluating vision science methods require robust and efficient tools for assessing their performance in various real-world scenarios. This study presents a novel virtual reality (VR) simulation tool that simulates real-world optical methods while giving high experimental control to the experiment. The tool incorporates an experiment controller, to smoothly and easily handle multiple conditions, a generic eye-tracking controller, that works with most common VR eye-trackers, a configurable defocus simulator, and a generic VR questionnaire loader to assess participants' behavior in virtual reality. This VR-based simulation tool bridges the gap between theoretical and applied research on new optical methods, corrections, and therapies. It enables vision scientists to increase their research tools with a robust, realistic, and fast research environment.
\end{abstract}
\begin{document}

\flushbottom
\maketitle

\thispagestyle{empty}

\section*{Introduction}
Vision impairments represent a significant global health concern, impacting millions of individuals. Refractive errors, encompassing conditions like myopia, hyperopia, astigmatism, and presbyopia, impair the eye's ability to focus light directly onto the retina, leading to blurred vision. Current estimates about the trend of myopia stay at 3.4 billion people, or half the world's population, by 2050 \cite{george2023myopia}. Beyond refractive errors, cataracts pose a substantial challenge, clouding the eye's lens and progressively impairing vision. Age-related macular degeneration (AMD) is another critical issue characterized by the loss of central vision and significantly hampering daily activities like reading and facial recognition. Its prevalence is estimated at 7.7 million people worldwide \cite{jiang2023trends}. Glaucoma, encompassing a group of conditions that damage the optic nerve, often due to high intraocular pressure, poses a severe risk of vision loss and blindness if left untreated. It is estimated to affect 79.6 million people worldwide and remains the leading cause of irreversible blindness \cite{quigley2006number}. Additionally, developmental disorders like amblyopia (Lazy Eye) and strabismus (Crossed Eyes) can lead to lifelong vision impairment if not addressed in their early stages. Collectively, these conditions represent the most prevalent vision impairments known to medical science.
Understanding the etiology of these impairments and discovering effective correction, treatment, or healing methods is imperative. However, transitioning from conceptual research to publicly available solutions is time-intensive.

In Paul Hibbard's book \cite{hibbard2023virtual}, the connection between VR and vision science and how we can use VR to improve vision science research is explained in detail. Hibbard shows the importance and strength of combining them. Specifically for vision science, no foundational tool helps scientists build their experiments. While there are approaches to using VR for vision science, they fail to support new experiments and most often have a specific purpose. BlueVR is specifically implemented to raise awareness for color blindness \cite{you2023bluevr}, Albrecht et al. \cite{albrecht2023mopedt} built a headset to augment peripheral vision exclusively called MoPedT. Their display toolkit can be built with off-the-shelf products, allowing designers and researchers to develop new peripheral vision interaction and visualization techniques. While MoPedT has a high external validity as it augments reality, it might help in real-life situations. However, developing new optical methods heavily relies on high experimental control to understand the dynamics under controlled circumstances, which cannot be achieved in real-life augmentations. Another closer related paper from Barbieri et al. \cite{barbieri2023realter} describes Realter, a reality-altering simulator for raising awareness of low-vision diseases like age-related macular degeneration, glaucoma, and hemianopsia. Their main focus is on simulating specific diseases rather than creating a foundational architecture for other researchers. Realter works as an extended reality device to simulate the diseases mentioned above while allowing the collection of eye- and head-tracking data from an HTC Vive Pro Eye with an integrated Tobii eye-tracker. Thus, it is restricted to specific devices and diseases. Like Albrecht et al., Barbieri et al. focused on application in the real world rather than on high experimental control in virtual reality. As such, both devices are great tools for applied research in the field to simulate specific diseases and conditions and collect data about human behavior but are restricted to real scenes, therefore missing the possibilities of virtual reality. Next to mainly disease simulation-focused works, other works aim to provide a simulator to improve specific ophthalmic conditions or hardware.
One exemplary simulator is described in the work of Niessner et al. \cite{Niessner2012}. They simulate eye accommodation in their simulation of human vision using different surfaces. They show two ways of simulating defocus. The progressive distributed ray tracing of eye lenses is an accurate approach to guide lens manufacturers during the design process. The second approach is implemented to give customers a real-time impression of spectacles in an eye shop. In their system, they additionally visualize the refractive power and astigmatism that allow a quality assessment of special-purpose lenses for sports or reading. Another work on VR simulators for vision science was published by Barbero et al. \cite{Barbero2017}. Their work shows a procedure to simulate real-world scenes through spectacle lenses. They aim to anticipate the effects of image transformation by optical defects found in spectacle lenses. They focus on blur, distortion, and chromatic aberration to evaluate ophthalmic lenses in a VR simulator.

Inspired by the works as mentioned earlier, we have developed an innovative simulation tool to expedite the development, robustness, and testing ease of new optical solutions for vision correction.  Our approach can be seen as a foundation for works like Niessner et al. to speed up such developments in a broader range. This tool combines an accurate optical method simulation and an evaluation procedure in a controlled environment, allowing for testing and refinement before implementation in physical devices. As a virtual reality simulator, VisionaryVR can realistically simulate blur caused by optical aberrations and possibly glaucoma and AMD. While VR allows a free, natural, and explorative behavior in evaluating different conditions, it simultaneously supports a high experimental control to investigate several types of vision impairment and environmental interactions scientifically. The simulation of MD or glaucoma is crucial as it can help understand affected people's behavior and create proper treatments or therapies (as shown in \cite{sipatchin2022application}). 
Our tool is anchored in an experiment controller, providing a foundational architecture for a virtual reality simulator. This simulator enables experimenters to seamlessly integrate independently built scenes applicable to various conditions in randomized or sequential order. Core components of this tool include an eye-tracking controller compatible with leading VR eye trackers, a defocus simulator capable of emulating different levels of defocus (and, therefore, simulating refractive errors), and an integrated questionnaire scene loader for direct VR responses. 
VisionaryVR is a foundational simulator for eye-tracking, focus simulation, and VR questionnaire experiments under multiple conditions. Its user-friendly and expansible design makes VisionaryVR a pivotal component in developing custom simulators for various sensors and actuators. Through this innovative tool, we aim to bridge the gap between theoretical research and practical application, paving the way for advancements in optical health and vision correction technologies. A significant application of VisionaryVR is as a testing environment for devices based on optoelectronic lenses, which are under active research. Considering hardware tuning mechanisms and natural interaction modes, VisionaryVR's modular structure facilitates comparing and evaluating different methods or conditions before their time-consuming implementation in hardware prototypes.  \\
The paper proposes a scene-based evaluation framework that recreates natural scenarios to address the limitations of current approaches and explore the potential of virtual reality simulation for the research of ophthalmic solutions. This framework aims to comprehensively assess visual performance and convenience, simulating individuals' real-life situations. By leveraging VR technology and the simulation tool, the paper aims to create an immersive framework that closely mimics these scenarios, facilitating precise measurement and analysis of the performance and convenience of optical methods or therapies. The objective of this work is to advance vision science research. By providing a realistic and controlled environment, the tool can potentially enhance real-world applications in various domains, including presbyopia correction, depth estimation, eye-tracking, and intention prediction. It can potentially revolutionize solutions for optical vision correction, enhancing the visual experience and quality of life for individuals across different fields, such as augmented reality, virtual reality, 3D imaging, and human-computer interaction. Through its comprehensive evaluation of a broad range of applications in optical systems research, the paper emphasizes the potential of the simulation tool in advancing the field. This tool aims to empower individuals with enhanced visual capabilities and enhance their everyday experiences.

\section{Methods}

\begin{figure}[h]
    \centering
    \includegraphics[width=0.5\textwidth]{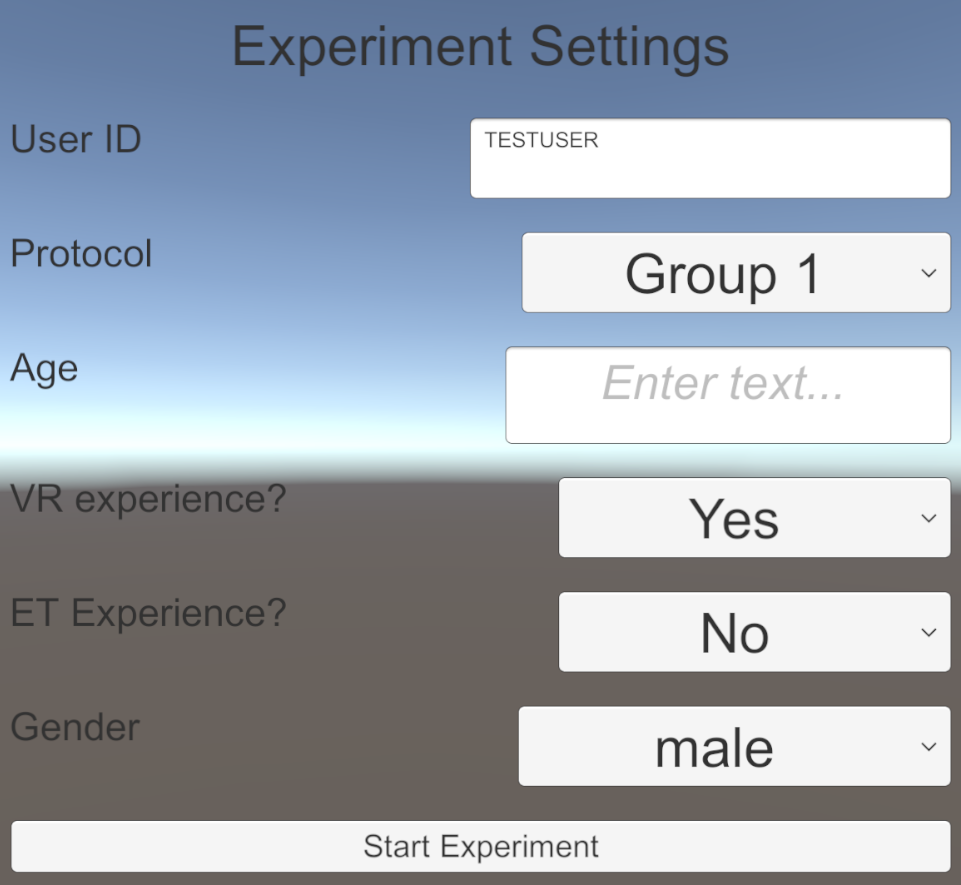}
     \caption{For every experiment, subject tracking is an essential part. The input mask is adaptable to incorporate the specific needs of the experimenter.}
    \label{fig:mask}
\end{figure}

\subsection{Experiment control}

\begin{figure}[h]
    \centering
       \begin{subfigure}[t]{0.49\textwidth}
        \includegraphics[width=\textwidth]{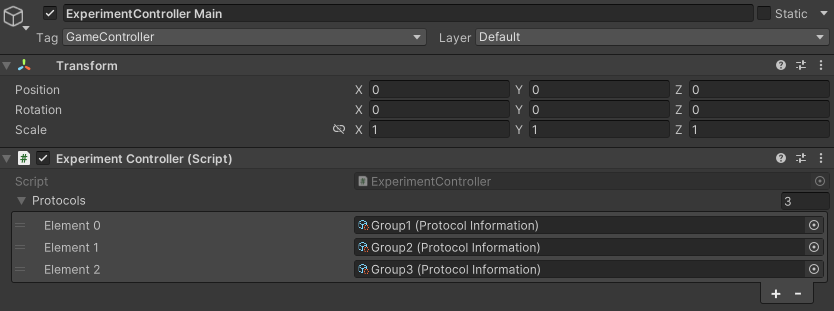}
        \caption{}
    \end{subfigure}
     \hfill
    \begin{subfigure}[t]{0.49\textwidth}
        \includegraphics[width=\textwidth]{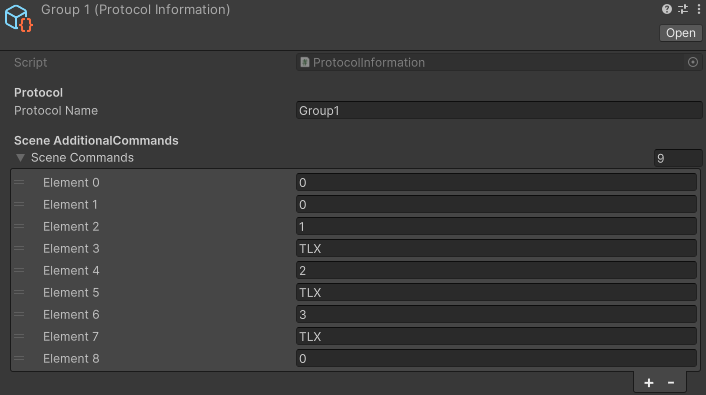}
        \caption{}
    \end{subfigure}
    \hfill
    \begin{subfigure}[t]{0.49\textwidth}
        \includegraphics[width=\textwidth]{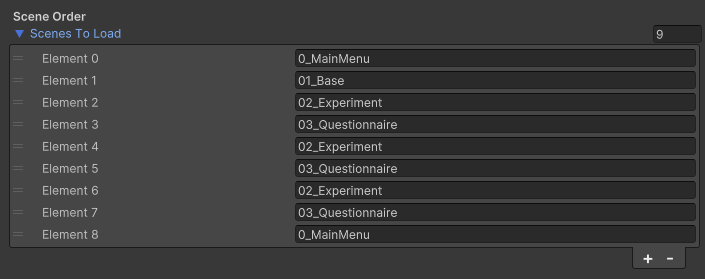}
        \caption{}
    \end{subfigure}      
    \caption{ (A) The heart of VisionaryVR is the experiment controller that controls the main thread by loading, unloading, and parameterizing each scene and synchronizing events and global objects across scenes. (B) Each protocol object can be added to a list in the experiment controller. This list is public and can be seen in the main menu dropdown. The experimenter can select which run configuration to choose.}
    \label{fig:protocols}

\end{figure}

When starting VisionaryVR, a user mask (Figure \ref{fig:mask} is showing up for the experimenter. The user name and basic demographic questions can be put in this mask. The content is adjustable to the experiment's needs. When starting the experiment, VisionaryVR creates a user folder inside the VisionaryVR global folder (which is changeable in the inspector). All current run data from all conditions and questionnaires are saved in this user folder. If the folder already exists, an incremented number is attached. The experiment controller is a scene loader that controls the game objects that must be loaded anytime throughout the experiment. Its main task is to take care of the order of the scenes and provides functionality like repetition, restarting, forwarding, or backwarding to another scene. The experiment controller manages the whole experiment by starting scenes and sending events to all listeners. Each scene can be seen as either a condition scene or a questionnaire scene. The experiment controller can parameterize all of them.  An experiment can also contain different run configurations represented by protocols (Figure \ref{fig:protocols}). In a protocol, each scene that must be shown is added to the scene order list. These scenes can be developed independently and must only be added to a protocol. The experimenter can pass a parameter for each scene in the scene order list. This parameter can be used for generic purposes. For a questionnaire scene, an abbreviation of the questionnaire to load is necessary for the loader to load the canvas dynamically with the correct questions.
In Figure \ref{fig:protocols} (A), the protocol "Group 1" contains the main menu (see Figure \ref{fig:mask}), a baseline scene, an experiment scene, and a questionnaire scene. If most of the functionality of a scene can be reused, a scene can be loaded multiple times with different parameters that can be used to select the condition in the same scene but with different methods. 

\subsection{Eye-Tracking}

\begin{figure}[h]
    \centering
    \includegraphics[width=0.8\textwidth]{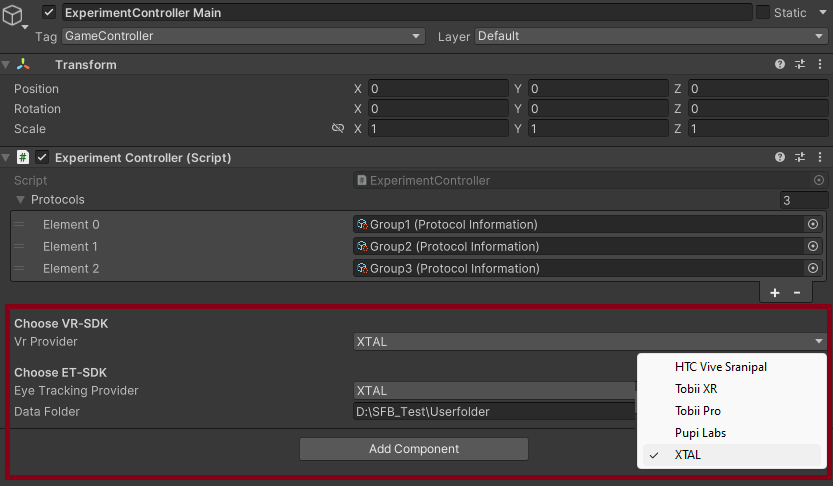}
    \caption{The figure shows the inspector view of the experiment controller game object. The settings for ZERO are shown in the red box. Next to the VR Provider (VR glasses), the experimenter can decide which eye-tracking provider API to use. The currently implemented APIs are shown in a dropdown menu on the right. If the gaze files should be saved in a specific location, this can be changed in the data folder input.}
    \label{fig:zero}
\end{figure}

Additionally, integrating ZERO \cite{hosp2023zero}, an open-source eye-tracking controller interface, enhances the simulation tool's capabilities. ZERO provides a standardized interface for various eye-tracking devices in virtual reality (for an example of the inspector view, see Figure \ref{fig:zero}, simplifying the implementation and usage within the VR environment. This integration enables seamless eye-tracking data collection and analysis, further enhancing the tool's potential for studying human gaze behavior and interactions with optical systems. By evaluating the participants' gaze behavior, we can assess the performance of the gaze-based interactions.
ZERO can directly implement the basic functionality needed for eye-tracking studies. Next to the easy usage of the most common eye-tracking APIs, the user has a certain level of device-independent control and can quickly start and stop the eye-tracker and save the gaze files to a specified user folder location. For most studies, it can be used out of the box. If further refinement is needed, the open-source software can be adapted to one's specific needs, as the interface is built highly modular and intended to be expandable by adding new eye-tracker devices via a config file. The gaze signal is saved in a generic file format that involves all essential information needed from the eye-tracker and some unique data of specific vendors. ZERO provides calls to the interface to customize the behavior. Experimenters can start and stop the eye-tracker device and the signal sampling thread separately. If the device supports a position or eye-tracking calibration, it can be called via the ZERO interface.

\subsection{Optical simulation}

\begin{figure}[h]
    \centering
    \begin{subfigure}[b]{0.32\textwidth}
        \includegraphics[width=\textwidth]{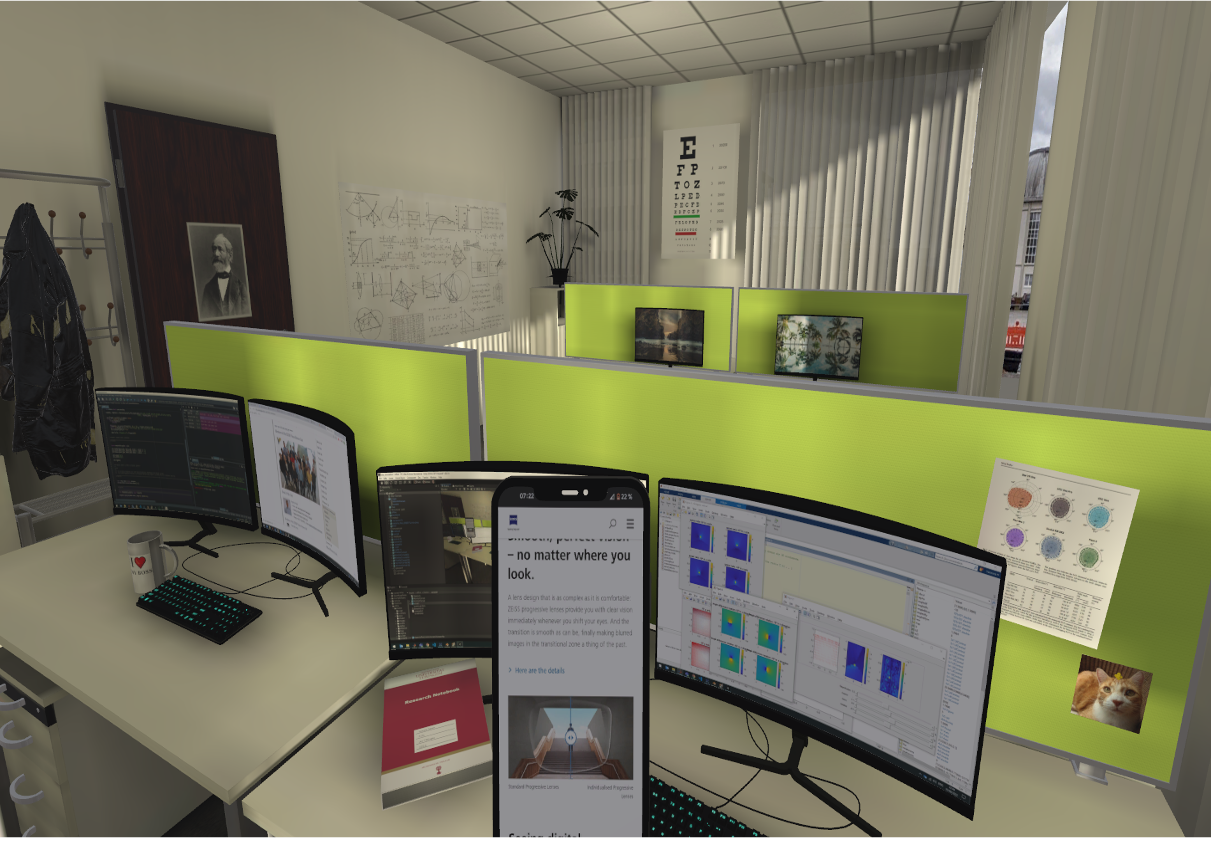}
        \caption{Original scene}
        \label{fig:scene}
    \end{subfigure}
    \hfill
    \begin{subfigure}[b]{0.32\textwidth}
        \includegraphics[width=\textwidth]{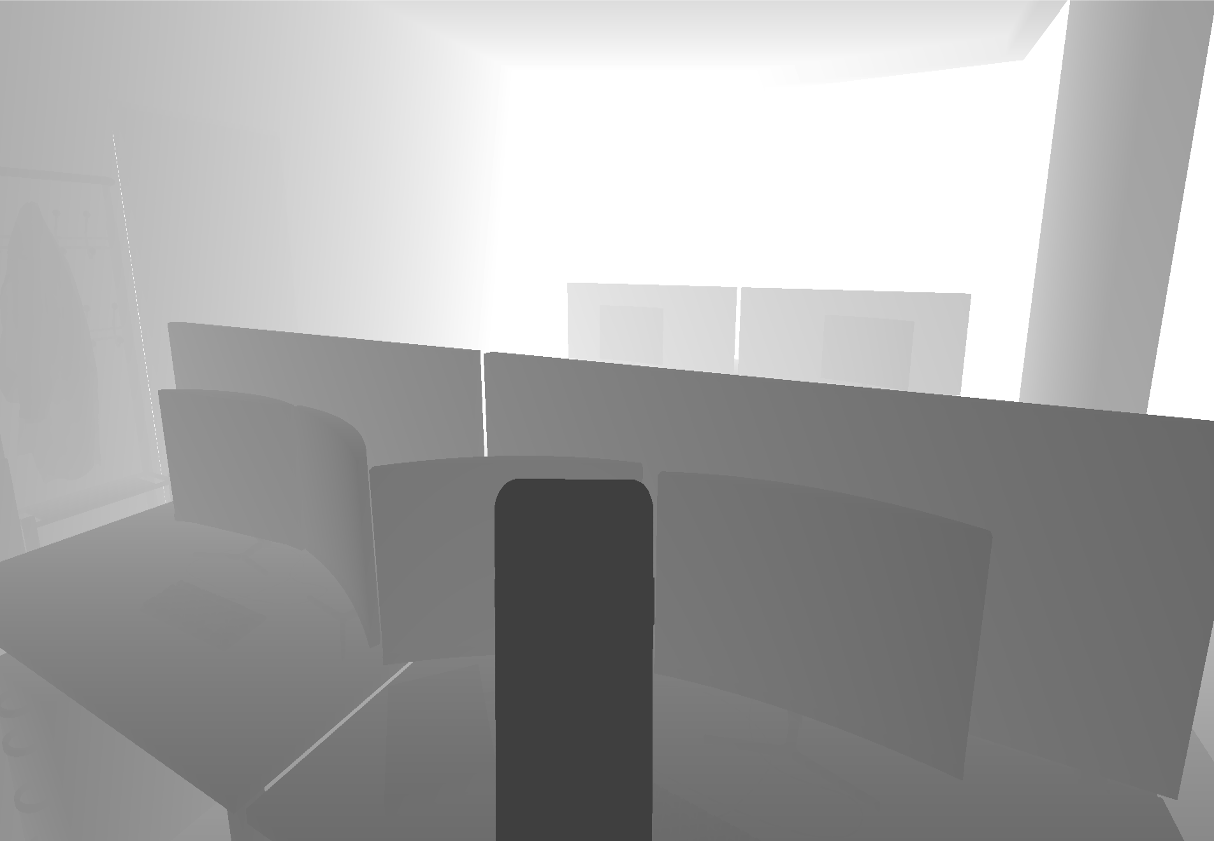}
        \caption{Depth map} 
        \label{fig:depthmap}
    \end{subfigure}
    \hfill
    \begin{subfigure}[b]{0.32\textwidth}
        \includegraphics[width=\textwidth]{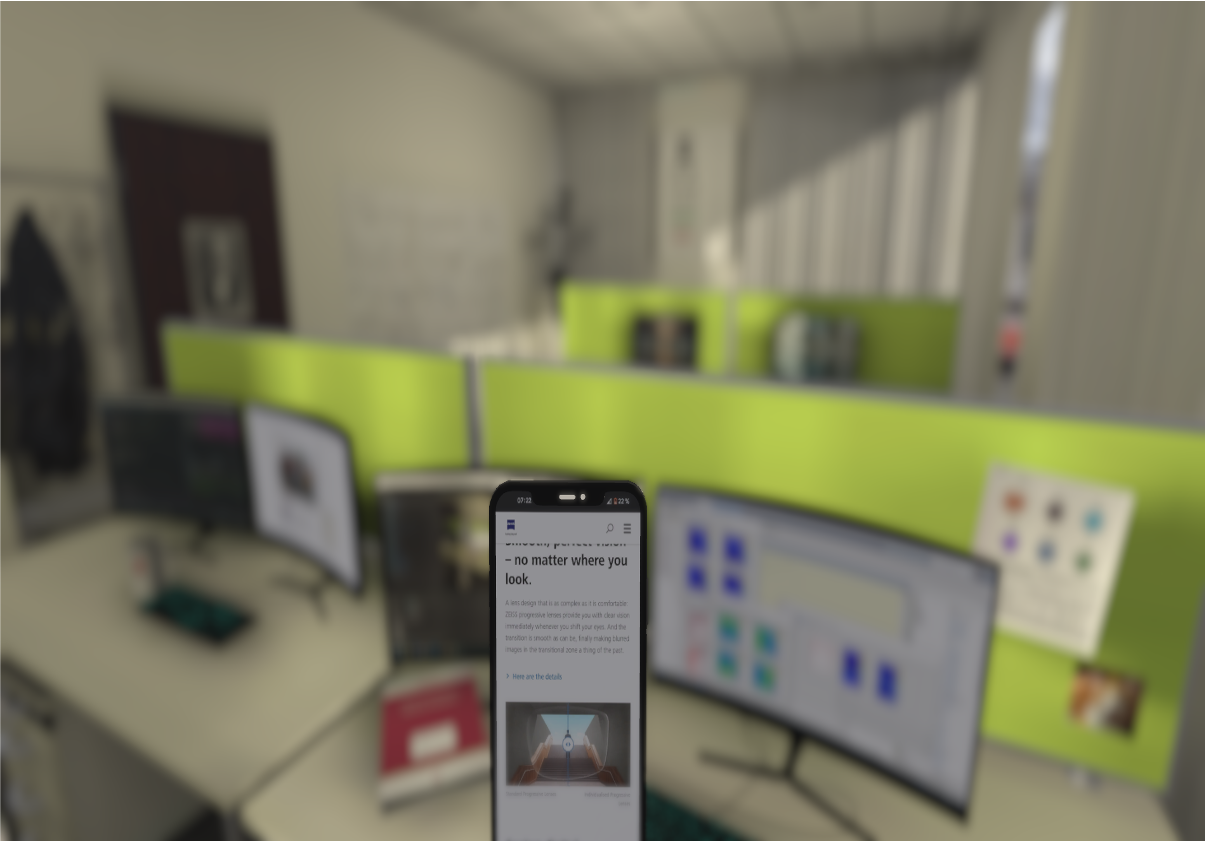}
        \caption{Defocused scene} 
        \label{fig:resultingscene}
        \end{subfigure}
        
    \caption{(A) A common office scene rendered in virtual reality. Several objects are shown in different depth planes. Ranging from close (smartphone) to far ( e.g. whiteboards). (B) The corresponding depth map of the same scene. During rendering, the depth map can be accessed in the shader, allowing the calculation of a location-dependent blur size. For the simulation of autofocals, the depth information is also used to simulate different control algorithms that determine the target focus distance for the user's current gaze point depending on the depth distribution in the scene (C) Simulated vision with an autofocal. The simulated lens is tuned to the distance of the smartphone. The rest of the scene is blurred.}
\end{figure}

The VR simulation of optical aberrations is based on applying depth-dependent blur to the rendered image \cite{Sauer2022c}. Refractive errors, as well as different optical corrections, can be simulated realistically. The blur size, shape, and orientation are calculated dynamically and location-dependently, allowing the simulation of spatially varying optical power necessary for the simulation of progressive lenses. An important focus of the simulation is on autofocals, based on simulating temporally varying optical power. A controller for the simulated lens power is accessible via a script attached to a game object. Considering pupil size and the variable optical power of the simulated focus-tunable lens, the simulation calculates location-dependent blur size \cite{strasburger2018blur}. By manipulating the simulated power of the autofocal lens using arbitrary control algorithms, the simulation determines the focus distance, resulting in a natural depth of field effect where objects near the current focus distance appear sharp. This simulation empowers researchers to assess the impact of autofocal defocus on visual performance and comfort under various control mechanisms. Moreover, the simulation tool considers other relevant factors that impact focus performance, such as pupil size, lighting conditions, and environmental factors. The principle of the exemplary blur simulation is based on shaders in the game engine and, therefore, easily applicable to other effects.
By incorporating these factors into the simulation, researchers can evaluate how different optical systems and control algorithms perform under varying conditions, providing valuable insights for optimizing their performance in highly natural, controlled environments.
An example scene, the depth map, and the resulting view are shown in Figures \ref{fig:scene}, \ref{fig:depthmap}, and \ref{fig:resultingscene}. This approach ensured that only objects near the distance fitting the current optical power appeared sharp.

\begin{figure}[h]
    \centering
    \begin{subfigure}[t]{0.49\textwidth}
        \includegraphics[width=\textwidth]{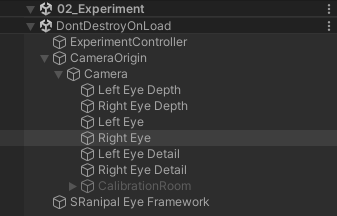}
        \label{fig:def_obj_cam}
    \end{subfigure}
    \hfill
    \begin{subfigure}[t]{0.49\textwidth}
        \includegraphics[width=\textwidth]{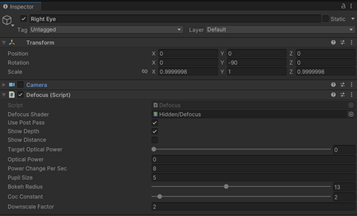}
        \label{fig:def_inspector}
    \end{subfigure}        
    \caption{ (A) Dynamically attached defocus script to the main camera (here Right Eye and Left Eye Camera). (B) The Inspector's live view of the autofocal controller shows the settings and configurations of the autofocal controller of the selected camera. The main purpose is to give the experimenter a live view of the current state of the autofocal controller (e.g., which depth level is currently in focus).}
\end{figure}

\subsection{Dynamic vision task}

\begin{figure}[h]
\centering
\includegraphics[width=\textwidth]{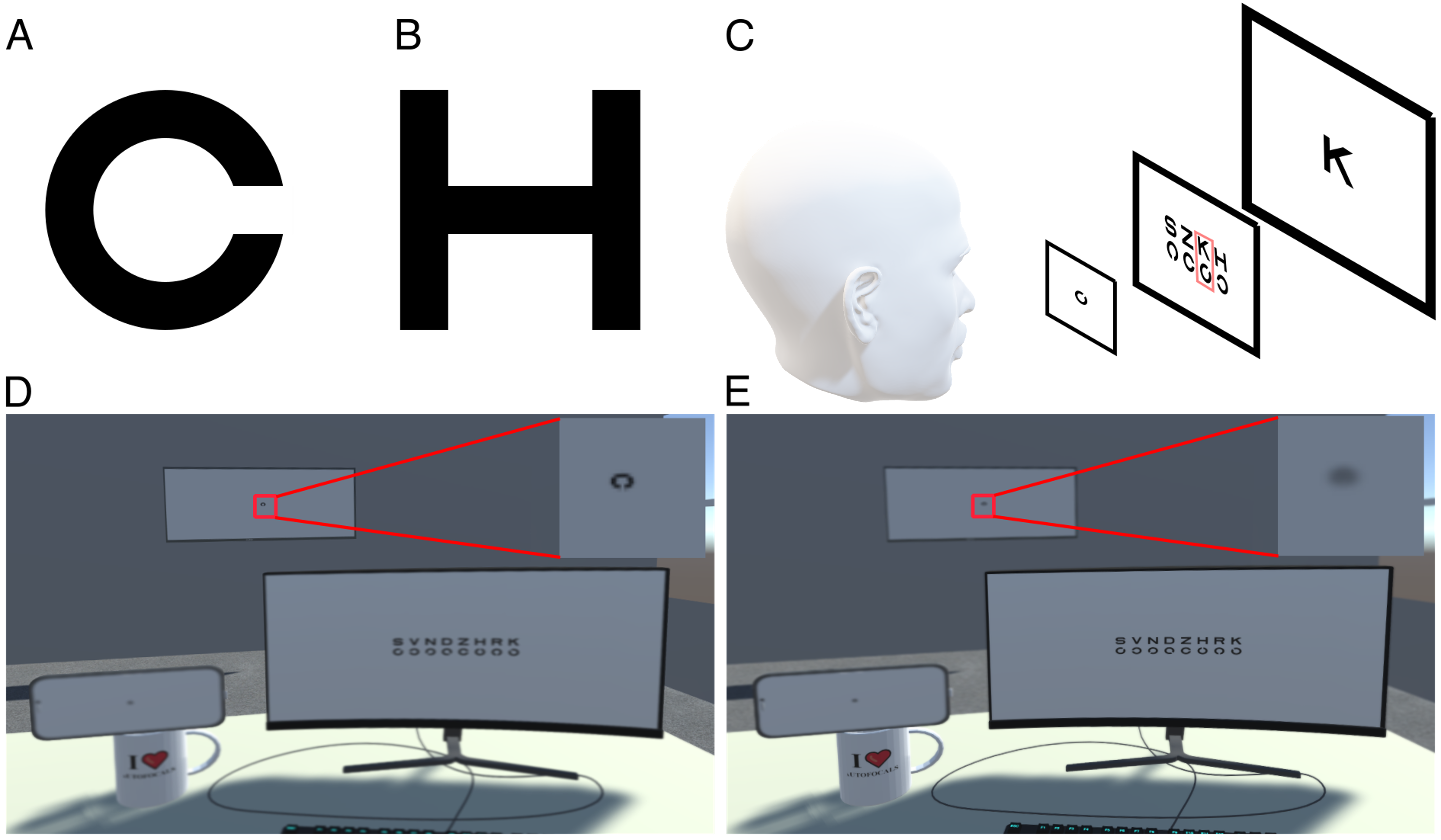}
\caption{Components of the matching task with exemplary focus and defocus. A) Landolt ring is the first type of stimulus used for the matching task. The opening gap can be oriented in 8 different directions. B) One of the 8 used Sloan letters is presented on a second screen. C) Stimuli are presented on three different screens at different distances. The closest screen represents a smartphone at a reading distance \SI{30}{\cm}. At an intermediate distance of \SI{1}{m}, there is a screen representing a computer display. The visual performance for far vision is tested at \SI{6}{m} distance. One of the screens shows a Landolt ring of random orientation, and a second screen shows a Sloan letter. A table of both stimuli types is displayed on the third screen. The task for the subjects is to answer if the two single stimuli are in the same table column on the third screen, corresponding to a match. This requires a focus shift between all three viewing distances. D) Screenshot of the matching task in the virtual environment. Defocus blur is simulated considering the virtual focus controller's local depth and dynamically set focus distance. In this example, the focus distance is set to the far screen. E) The same scene with focus distance set for the intermediate screen. The Landolt stimulus on the far screen is blurred and out of focus.}
\label{fig:matching_task} 
\end{figure}

Visual performance is usually evaluated for a fixed distance only. For dynamically evaluating performance, we designed a matching task with multiple viewing distances to simplify the VR scenarios into psychophysical paradigms that could be easily repeated. This task requires participants to shift their focus between screens displaying stimuli at various distances, simulating real-world situations involving dynamic gaze changes. Our simulator recreates a typical office environment with different viewing distances. Subjects are presented with stimuli on multiple virtual screens, requiring them to change their gaze dynamically. The VR simulation lets us dynamically set the virtual focus distance by considering depth information from the 3D scene, allowing for calculating realistic depth-dependent blur. By implementing a matching task that compares visual stimuli on different screens, we can objectively quantify visual performance without restricting the subjects' natural gaze and head-movement behavior. This setup accurately represents real-world scenarios and allows for the precise measurement of visual performance. The stimuli on the screens are normally distributed and shown at the center or the corner of the screen to incorporate complex cases where a stimulus is close to the border of two distinct depths. The developed task with multiple viewing distances compares visual performance and convenience in a realistic everyday scenario with dynamic gaze changes. Stimuli were shown on three different screens at three different distances: a smartphone at 30 cm, a computer display at 1 m, and a far-distance TV screen at 6 m. This arrangement elicits many dynamic changes in viewing distance, as subjects must perform a matching task with stimuli placed on all three screens. Landolt rings, and Sloan letters are used as standardized optotypes for visual acuity testing to reduce task complexity. Subjects must compare if the given combination of Landolt ring and Sloan letter appeared in the same table column displayed on the third screen. Using a combination of two different types of stimuli, we reduce task complexity while maintaining sensitivity to defocus blur. Randomizing the screen displaying the table prevented subjects from following a fixed order of fixating the screens, resulting in dynamic and better comparable gaze changes to natural behavior.  A simple visualization of the task in two different focus depths can be seen in Figure \ref{fig:matching_task}.

\subsection{Questionnaires}

To ensure the effectiveness and suitability of optical systems for everyday use, it is essential to replicate the natural behavior of individuals in real-world scenarios. Researchers can optimize the design and control mechanisms to meet users' specific needs and preferences by studying how individuals interact with and adapt to optical methods. Psychophysical paradigms and behavioral quantifiers play a vital role in understanding visual perception and performance aspects, quantifying the benefits and limitations of new methods, and comparing their performance against traditional methods. Behavioral quantifiers also provide objective user experience and satisfaction measures, enabling researchers to gain valuable insights into users' preferences and inform future improvements and customization of these devices. The VR questionnaire loader is implemented in a separate scene and loaded when mentioned in the protocol. Which questions to load can be decided in the choice of the transfer parameters in the experiment controller. The questionnaire to be used can be specified for each scene separately. As long as a .json file with the questions is stored, it can be loaded. The file is dynamically read out at runtime and displayed in virtual reality. The user can operate the questions via ordinary VR controllers. The data is stored directly in the user folder in a subfolder containing the scene name.

\begin{figure}[h]
    \centering
    \includegraphics[width=0.9\textwidth]{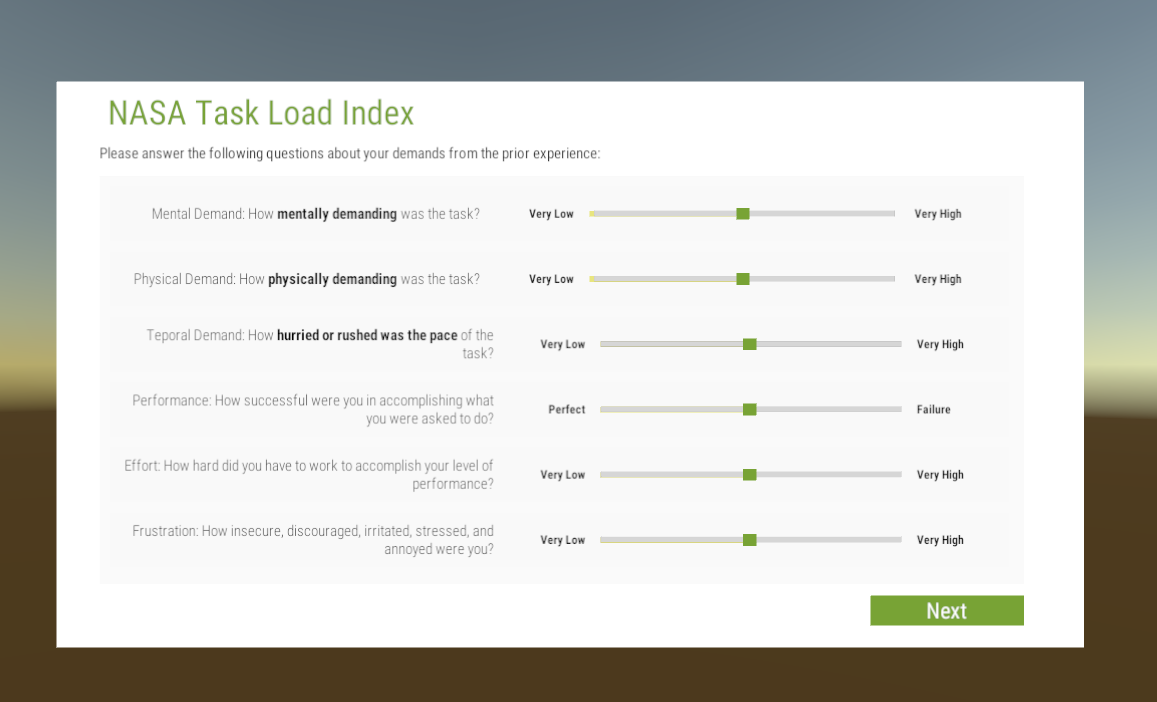}
    \label{fig:questionnaire}
    \caption{ Provided as an abbreviation in the experiment controller (e.g., TLX), a questionnaire scene is loaded that is shown in VR for the subject. Depending on the abbreviation, the questions of the correct questionnaire are loaded dynamically into the canvas. Interaction takes place using VR controllers. The experimenter can see the same view on the screen during the whole scene.}
\end{figure}

\section*{Discussion}


This paper highlights the potential of VisionaryVR, a VR-based simulation tool for testing, evaluating, and optimizing optical systems. The system allows experimenters like designers or researchers to create their own experiments for vision research. VisionaryVR acts as a foundation architecture to support those projects. Next to its easy and parameterizable scene handling, it already supports several common eye trackers out of the box, has a focus controller to change the focus of eye cameras to simulate different refraction errors, and includes a questionnaire loader. All functions are directly usable or adjustable to one's specific needs. Published under a user-friendly CC 4.0 license, it is open-source and made for the community. The simulator could further enable the assessment of other solutions for eye diseases such as cataracts, glaucoma, and AMD with better configurability and even more opportunities for scientists with low interest in developing the environment independently.\\
The simulation tool used in this study has broader implications beyond evaluating optical methods. One significant application is depth estimation. Accurate depth estimation is crucial in various domains, such as augmented reality, virtual reality, and robotics. The simulation tool can be leveraged to evaluate and refine depth estimation algorithms, providing a controlled environment to validate their performance and optimize their accuracy. This can lead to advancements in in-depth estimation techniques, ultimately improving the realism and effectiveness of applications relying on accurate depth perception. Eye-tracking systems also stand to benefit from the simulation tool. Eye tracking is critical in various fields, including human-computer interaction, usability testing, and healthcare. By simulating different eye movement patterns and gaze behaviors, the tool can aid in developing and evaluating eye-tracking algorithms, facilitating improvements in accuracy, robustness, and tracking speed. This, in turn, can enhance the performance and usability of eye-tracking systems across a wide range of applications. Additionally, the simulation tool can contribute to developing intention prediction models. Anticipating human intention is essential in areas such as autonomous vehicles, human-robot interaction, and assistive technologies. By simulating various scenarios and user behaviors, the tool can assist in refining and validating intention prediction models, improving their accuracy and reliability. This can have significant implications for safety, efficiency, and overall user experience in contexts where accurate intention prediction is critical. \\
While providing a controlled and repeatable environment, the VR simulation may not fully replicate the complexities of real-world accommodation and lens-tuning interactions. Real-world implementation of optical methods would involve various environmental factors, dynamic focus adjustments, and individual user characteristics that may not be fully captured in a simulated environment. Therefore, choosing the optical method to work on in the simulator must be done thoroughly. For instance, if vergence movements in combination with accommodation are included, experimenters have to consider the existing limitations of VR, such as the vergence-accommodation conflict, in their analysis. In such cases, the results obtained in the VR simulation may not be directly translated to real-world scenarios. They could affect the generalizability of the findings to real-world implementation.
The value of the simulation tool utilized in this study is evident in its ability to evaluate and compare different optical methods within a controlled and repeatable environment. By fostering interdisciplinary collaborations, we further advance vision science research and promote the development of more effective and convenient optical solutions. 

Overall, the utilization of the simulation tool contributes to the advancement of vision science research. By addressing the limitations and refining control mechanisms, researchers and developers can work towards providing individuals with vision impairments with more efficient, convenient, and effective visual solutions.

\section{Data \& Code Availability}

The simulation tool's source code and a unity package are publicly available at https://github.com/zvsl-ai/VisionaryVR.

\bibliography{main}

\section*{Acknowledgements}

The German Research Foundation (DFG) generously supported this research under SFB 1233, Robust Vision: Inference Principles and Neural Mechanisms, TP TRA, with project number 276693517.

\section*{Author contributions statement}
BWH contributed to the development, data collection, analysis, writing, and hypothesis formulation. YS contributed to the development, report, and hypothesis formulation. RA contributed to writing and hypothesis formulation. SW contributed to writing, hypothesis formulation, and project funding. The authors declare no competing interests.


\end{document}


\appendix

\section{Questionnaire Evaluation: Manual}
\begin{table}[htbp]
    \centering
    \begin{tabular}{ccccccc}
        \toprule
        Subject & Mental Demand & Physical Demand & Temporal Demand & Performance & Effort & Frustration \\
        \midrule
        1 & 0 & 0 & 10 & 20 & 60 & 0 \\
        2 & 20 & 10 & 10 & 10 & 40 & 70 \\
        3 & 70 & 60 & 50 & 20 & 60 & 60 \\
        4 & 90 & 90 & 0 & 30 & 80 & 30 \\
        5 & 40 & 0 & 40 & 20 & 50 & 20 \\
        6 & 20 & 10 & 0 & 30 & 0 & 0 \\
        7 & 10 & 10 & 20 & 10 & 10 & 10 \\
        8 & 40 & 30 & 10 & 10 & 40 & 20 \\
        9 & 40 & 30 & 30 & 50 & 50 & 50 \\
        10 & 80 & 80 & 40 & 40 & 70 & 40 \\
        11 & 50 & 100 & 0 & 40 & 100 & 101 \\
        12 & 50 & 50 & 30 & 10 & 30 & 30 \\
        13 & 60 & 50 & 40 & 30 & 40 & 60 \\
        14 & 90 & 60 & 10 & 0 & 10 & 10 \\
        15 & 60 & 30 & 40 & 10 & 60 & 40 \\
        16 & 45 & 50 & 50 & 50 & 50 & 50 \\
        17 & 20 & 30 & 50 & 10 & 60 & 20 \\
        18 & 70 & 10 & 10 & 30 & 30 & 10 \\
        19 & 70 & 70 & 40 & 40 & 60 & 30 \\
        20 & 60 & 60 & 40 & 40 & 60 & 60 \\
        \bottomrule
    \end{tabular}
    \caption{The table shows all subjects with their rating on the NASA TLX questionnaire for the manual condition. The single columns describe different dimensions of cognitive demand. Answer rating ranges from 0 = not agree/low demand to 100 = agree/ high demand.}
    \label{tbl:manual_all_questions_all_subjects}
\end{table}

\section{Questionnaire Evaluation: Gaze}
\begin{table}[htbp]
    \centering
    \begin{tabular}{ccccccc}
        \toprule
        Subject & Mental Demand & Physical Demand & Temporal Demand & Performance & Effort & Frustration \\
        \midrule
        1 & 0 & 60 & 10 & 30 & 80 & 60 \\
        2 & 40 & 60 & 10 & 30 & 70 & 80 \\
        3 & 70 & 70 & 50 & 50 & 60 & 60 \\
        4 & 0 & 100 & 0 & 10 & 60 & 80 \\
        5 & 40 & 10 & 40 & 30 & 80 & 70 \\
        6 & 10 & 10 & 0 & 10 & 0 & 0 \\
        7 & 30 & 30 & 40 & 20 & 80 & 20 \\
        8 & 90 & 90 & 10 & 40 & 90 & 90 \\
        9 & 40 & 60 & 20 & 20 & 70 & 30 \\
        10 & 80 & 60 & 30 & 20 & 60 & 60 \\
        11 & 30 & 30 & 50 & 10 & 60 & 70 \\
        12 & 70 & 100 & 0 & 30 & 100 & 101 \\
        13 & 70 & 70 & 50 & 20 & 70 & 50 \\
        14 & 70 & 40 & 40 & 40 & 60 & 70 \\
        15 & 70 & 60 & 10 & 10 & 20 & 20 \\
        16 & 70 & 60 & 50 & 20 & 70 & 50 \\
        17 & 64 & 29 & 51 & 26 & 73 & 50 \\
        18 & 20 & 80 & 70 & 10 & 60 & 30 \\
        19 & 50 & 20 & 10 & 30 & 30 & 30 \\
        20 & 60 & 60 & 40 & 40 & 70 & 30 \\
        \bottomrule
    \end{tabular}
    \caption{Similar to table \ref{tbl:manual_all_questions_all_subjects}, the table shows all subjects with their rating on the NASA TLX questionnaire for the gaze condition. The single columns describe different dimensions of cognitive demand. Answer rating ranges from 0 = not agree/low demand to 100 = agree/ high demand.}
    \label{tbl:gaze_all_questions_all_subjects}
\end{table}